\documentclass{bmcart}

\usepackage{amsthm,amsmath}
\usepackage[utf8]{inputenc} 
\usepackage{graphicx}
\usepackage{times}
\usepackage{helvet}
\usepackage{courier}
\usepackage{graphicx}
\usepackage{subfigure}
\usepackage{indentfirst}
\usepackage{amsmath}
\usepackage{longtable}
\usepackage{multirow}

\startlocaldefs
\endlocaldefs

\begin{document}

\begin{frontmatter}

\begin{fmbox}
\dochead{Research}


\title{Revisit Lmser and its further development based on convolutional layers}


\author[
   addressref={aff1},                   
   email={huangwenjing@sjtu.edu.cn}   
]{\inits{WJ}\fnm{Wenjing} \snm{Huang}}
\author[
   addressref={aff1},
   email={tushikui@sjtu.edu.cn}
]{\inits{SK}\fnm{Shikui} \snm{Tu}}
\author[
   addressref={aff1,aff2},
   corref={aff1},    
   email={leixu@sjtu.edu.cn}
]{\inits{LX}\fnm{Lei} \snm{Xu}}


\address[id=aff1]{
  \orgname{Department of Computer Science and Engineering, and CMaCH Center, School of Electronic Information and Electrical Engineering, Shanghai Jiao Tong University}, 
  \street{800 Dongchuan Rd},                     %
  \city{Shanghai},                              
  \cny{China}                                    
}
\address[id=aff2]{%
  \orgname{Department of Computer Science and Engineering, The Chinese University of Hong Kong},
  \street{Shatin, N.T.},
  \city{Hong Kong},
  \cny{China}
}


\begin{artnotes}
\end{artnotes}

\end{fmbox}


\begin{abstractbox}

\begin{abstract} 
Proposed in 1991, Least Mean Square Error Reconstruction for self-organizing network, shortly Lmser, was a further development of the traditional auto-encoder (AE) by folding the architecture with respect to the central coding layer and thus leading to the features of symmetric weights and neurons, as well as jointly supervised and unsupervised learning. However, its advantages were only demonstrated in a one-hidden-layer implementation due to the lack of computing resources and big data at that time. In this paper, we revisit Lmser from the perspective of deep learning, develop Lmser network based on multiple convolutional layers, which is more suitable for image-related tasks, and confirm several Lmser functions with preliminary demonstrations on image recognition, reconstruction, association recall, and so on. Experiments demonstrate that Lmser indeed works as indicated in the original paper, and it has promising performance in various applications.
\end{abstract}




\end{abstractbox}
%

\end{frontmatter}



\section*{Introduction}


Early efforts on bidirectional multilayer networks can be traced back to 1980s. Three-layer networks, i.e., networks with only one hidden layer, were used to make auto-association to learn inner representations of observed signals \cite{Ballard1987,Bourlard1988AE}. In this framework, the network architecture is considered to be symmetric with the hidden layer as the central coding layer $Y$. The input patterns $X$ are mapped through the encoder part to the central coding layer, while the output $\hat{X}$ is constrained to reconstruct the input patterns via the decoder part to decode the internal representations back to the data space. The two parts, i.e., encoder $X\to Y$ and decoder $Y\to\hat{X}$, make an autoencoder (AE) network.

Least Mean Square Error Reconstruction (Lmser) self-organizing network was first proposed in 1991 \cite{xu1991LMSER,xu1993least}, and it is a further development of AE with favorable features. AE implements a direct cascading $X\to Y$ and $Y\to\hat{X}$ by a simple circle. The Lmser architecture is obtained from folding AE along the central coding layer $Y$, and then improves it into a distributed cascading by not only constraining $X\to Y$ and $Y\to\hat{X}$ to share the same architecture, but also using the same neurons for the two layers symmetrically paired between the encoder and decoder with respect to the central coding layer, and using the same connection weights for the bidirectional links between two consecutive layers along the directions of $X\to Y$ and $Y\to\hat{X}$. One neuron takes both roles in encoder and decoder, which can be regarded as adding shortcut connections between the paired neurons. Using the same connections weights for the bidirectional links, i.e., $A=W^T$ where $W$ is for the direction $X\to Y$ and $A$ is for $Y\to\hat{X}$, enables Lmser to approximate identity mapping simply through $WA=A^TA\approx I$ which holds exactly for an orthogonal matrix $A$, whereas AE implements a direct approximation of inverse of $X\to Y$ by $Y\to\hat{X}$ in a simple cycle. Due to the favorable architectural features, Lmser works in two phases, i.e., perception phase and learning phase. In perception phase, the signal propagation from two directions $X\to Y$ and $Y\to\hat{X}$ constitute a dynamic process which will approach equilibrium in a short term. If the reconstruction $\hat{X}$ is not close to $X$, then Lmser enters the learning phase to update the connection weights reduce the gap. Moreover, part of the central layer $Y$ can be used for label prediction $Y_L$ and thus supervised and unsupervised learning are made jointly by Lmser.

As discussed in \cite{xu1991LMSER}, Lmser potentially has many functions. However, due to the lack of powerful computing facility and big data at the time of $1990$'s, Lmser was implemented by computer simulations with only one hidden layer, and it was shown that a neuron in Lmser net behaved similar to a feature detector in the cortical field during learning \cite{xu1991LMSER}. In recent years, some features of Lmser were also adopted in the literature. For example, stacked restricted Boltzmann machines (RBMs) \cite{hinton2006fast} constrained the weight parameters to be symmetrically shared by the encoder and decoder network, while the neurons in U-Net \cite{ronneberger2015u} and deep RED-Net \cite{mao2016image} shared values between the two ends, using skip connections from neurons of one layer in the encoder to the neurons of another symmetrically corresponding layer in the decoder. Other related bidirectional deep learning models, such as Variational autoencoder (VAE) \cite{kingma2013auto} and Generative Adversarial Networks (GAN) \cite{goodfellow2014generative}, were also proposed and studied intensively. However, whether Lmser indeed works on deep network structures, effective for those potential functions as indicated in \cite{xu1991LMSER}, is still not systematically explored.

In this paper, we revisit Lmser by implementing it on a multi-layer network, and confirm that it works as indicated in \cite{xu1991LMSER,xu1993least}. We also extend the Lmser network from the fully-connected layers to the convolutional layers adapted from Convolutional Neural Network (CNN)\cite{lecun1989backpropagation}. Experiments confirm several potential functions of Lmser and demonstrate its promising performance in various applications. Our contributions are summarized as follows:

\begin{itemize}
	\item We revisit Lmser net by implementing it on multiple layers of neural networks. Our implementation can train the deep Lmser networks stably and effectively.
	\item In the original paper \cite{xu1991LMSER}, Lmser net was based on fully-connected layers, which is not good for image input. Thus, we extend to build the Lmser net from fully-connected layers to convolutional layers, to improve the performance on image-related tasks such as image recognition, generation, and so on.
	\item Experiments are conducted on image reconstruction, generation, associative recall, and classification, in comparisons with AE as a baseline. The results not only confirm that Lmser works as indicated in the original paper, but also demonstrate its promising performance in various applications.
\end{itemize}

\section*{Related Work}   \label{sec:related}



\subsection*{Networks with symmetrically weighted connections}

A stack of two-layer restricted Boltzmann machines (RBMs) with symmetrically weighted connections was used in \cite{hinton2006reducing} to pretrain an AE that has multiple hidden layers in a one-by-one-layer way. Such an AE is hard to optimize its weights jointly because the gradients vanishes as it propagates to early layers. The AE unrolled from a stack of pretrained RBMs worked well in dimensionality reduction.


\subsection*{Networks with symmetrically skip connections}

One recent typical example network architecture with symmetrically skip connections is the U-Net \cite{ronneberger2015u}. It consists of a contracting path as encoder and an expansive path as decoder, and both paths form a U-shaped architecture. The feature map from each of the layer of the contracting path was copied and concatenated with the symmetrically corresponding layer in the expansive path. Such skip connections can be regarded as a type of sharing the encoder neuron values with the correspondingly paired decoder neuron. Experiments in \cite{ronneberger2015u} demonstrated that U-Net worked very well for biomedical image segmentation. Such skip connections were also adopted in deep RED-Net \cite{mao2016image}. Different from U-Net, it directly adds old feature map with present top-down signal. Feature map must be in the same size, so it only add skip connections for specific layers. It has been shown in \cite{mao2016image} that deep RED-Net worked well in super-resolution image restoration. Moreover, with the appearance of ResNet\cite{he2016deep} and \cite{DenseNet2016}, deep neural networks with skip-connections become very popular and showed impressive performance in various applications.

\section*{A brief review of Lmser}   \label{sec:pre}

\begin{figure}
	\centering
	\includegraphics[width=6cm]{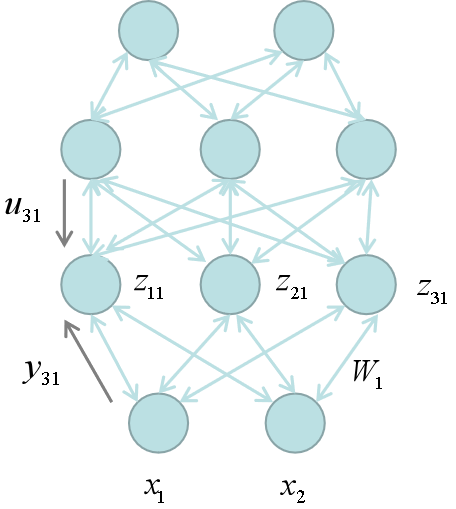}
	\caption{The architecture of multi-layer Lmser, where connection is bidirectional and symmetric, $u_{k}$ is top-down signal, $y_{k}$ is bottom-up signal, and each neuron is output as a sigmoid activation, i.e., $z_k=s(u_k+y_k)$.}
	\label{fig:LMSER}
\end{figure}

We briefly review the architecture and the two phase working mechanism of Lmser in the following. The Lmser self-organizing net was proposed in \cite{xu1991LMSER,xu1993least} based on the principle of Least Mean Square Error Reconstruction (LMSER) of an input pattern. As demonstrated in Figure \ref{fig:LMSER}, the architecture of Lmser net consists of either one layer or multiple layers. Each neuron receives both bottom-up signals from the lower layer and top-down signals from the upper layer. 

The Lmser net works in two phases, i.e., perception and learning. In the perception phase, the input pattern $X$ triggers the dynamic process by passing the signals up from the bottom layer, while simultaneously the signals in the upper layers will be passed down to the lower layers. The top-down signal to the input layer is regarded as reconstruction of the input pattern. It has been proved that the process will converge into an equilibrium state \cite{xu1991LMSER}. In the learning phase, the parameters are updated by minimizing the mean square error (MSE) between the input pattern and the reconstruction signals.

As given by Eq.(5a) in \cite{xu1993least}, the loss function for Lmser learning is as follows:
\begin{align}
	J &= \frac{1}{2}E({\|\overrightarrow{x}-W_1^T\overrightarrow{z_1}\|}^2)
	\label{eq:lmser-objective}
\end{align}
where $E(\cdot)$ denotes expectation by regarding $x$ as a random variable, $W_1$ is the weights of the first layer, $z_1=s(y_1+u_1)$ is the activity of the first layer neurons which receives both the bottom-up signals $y_1=W_1x$ from the input $x$ and the top-down reconstruction $u_1=W_1^Tz_2$ from the second layer $z_2$. In practice, one may use batch-way gradient on a batch of data $\{x_t\}_{t=1}^N$, or use stochastic gradient to update the parameters on one or a small set of data. As given by Eq.(6a)-Eq.(8b) in \cite{xu1993least}, the gradients to update the network parameters were restated below:
\begin{align}
	{\rm{ - }}\frac{{\partial J}}{{\partial {W_{pqk}}}}  = & \sum\limits_{{r_k} = 1}^{{n_k}} {{\omega _{{r_k}k}}\frac{{\partial {z_{{r_k}k}^t}}}{{\partial {W_{pqk}}}}}  + \sum\limits_{{r_{k - 1}} = 1}^{{n_{k - 1}}} {{\omega _{{r_{k - 1}}k - 1}}\frac{{\partial {z_{{r_{k - 1}}k - 1}^t}}}{{\partial {W_{pqk}}}}}
	\label{eq:chain-rule}
\end{align}
where
\begin{align}
	{\omega _{ik}} & =  - \frac{{\partial J}}{{\partial {z_{ik}^t}}} =  - \sum\limits_{{r_{k - 1}} = 1}^{{n_{k - 1}}} {\frac{{\partial J}}{{\partial {z_{{r_{k - 1}}k - 1}^t}}}} \frac{{\partial {z_{{r_{k - 1}}k - 1}^t}}}{{\partial {z_{ik}^t}}}  \nonumber  \\
	&=  - \sum\limits_{{r_{k - 1}} = 1}^{{n_{k - 1}}} {{\omega _{{r_{k - 1}}k - 1}}}s'_{ik-1}{W_{{r_{k - 1}}ik}^d}.
	\label{eq:chain-rule2}
\end{align}

\section*{Methods}   \label{sec:method}

\subsection*{Revisit Lmser and implement it on multiple fully-connected layers}

The Lmser learning rules proposed in \cite{xu1991LMSER} can be directly used for updating parameters in multiple layers. According to \cite{xu1991LMSER,xu1993least}, there could be many ways of implementation of deep Lmser learning over multiple hidden layers. In this paper, we present an effective implementation. In perception phase, it is hard to implement exact dynamic process. Instead, we propose an approximation to update the neuron activities in a layer-by-layer way, which is easy to implement and works well in practice. At the very beginning, a signal vector is placed at the input layer into the network, and then it will trigger the bottom-up signal propagation through the layers one by one. During this first bottom-up pass, at the $i$-th layer, the top-down signals from $(i+1)$-th layer to the $i$-th layer are initialized to be zero. After the first bottom-up pass, all neurons are given activity values. Then, the first top-down pass propagates the signals backwards and the neuron activation at the $i$-th layer is calculated as a sigmoid of the summation of the output from the $(i+1)$-th layer and the previous value of the $i$-th layer. In analogy to light reflections, we call such one bottom-up and one top-down pass as one reflection. In this way, bottom-up signal and top-down signal can be updated alternatively until they become stable. In practice, we use the Rectified Linear Units (ReLU) as the neuron activating function, and we find that one reflection (i.e., $k=1$) followed by learning phase works well for the whole Lmser learning.

\subsection*{CNN based Lmser learning}

The original Lmser net proposed in \cite{xu1991LMSER,xu1993least} takes vector input. When tackling with multichannel image with high resolution, one way is to reshape the image into vectors which is not good for extracting patterns in the images. For image-related tasks, we further develop Lmser network based on convolutional layers similar to Convolutional Neural Networks (CNN)\cite{lecun1989backpropagation}. Following original Lmser rules by Eq.(\ref{eq:chain-rule}), we built a CNN based Lmser net. 

In perception phase, we implement the dynamic process in a similar way to the reflection pass. There are multiple ways to fuse information from both bottom-up and top-down propagation. For example, U-Net concatenates bottom-up feature map and top-down feature map and performs convolution to achieve information fusion \cite{ronneberger2015u}. In this paper, we adopt the simple addition as in the original fully-connected layer based Lmser. For the bottom-up propagation, the standard convolution is implemented at each layer, while for the top-down propagation, the transposed convolution is used. The connection weights for the convolutional filters are shared by the two directions.



\subsection*{Jointly supervised and unsupervised Lmser learning}

In Lmser, each input pattern $X$ will be recognized by a label $Y_L$ output at the top layer, which plays the same role as a classifier. Thus, for the input with labels, Lmser can be implemented jointly both in an unsupervised manner to reduce the reconstruction error at the bottom layer and in a supervised way by minimizing the label loss between the predicted label $Y_L$ and the true one. The reconstruction error will push the network to do self-organizing, which helps network to learn structure information of input and facilitates concept abstracting and formation at the top domain. The self-organizing learning directed by reconstruction error plays the role of regularization, which help network to prevent over fitting and make classification be more robust such as defense against adversarial attacks.

In perception phase, the implementation of dynamic process is the same as the original Lmser, but with two parts computed at the top layer, i.e., one for predicted labels and the other as input to the decoder for top-down reconstruction. In learning phase, the loss function includes two terms for reconstruction and classification separately:
\begin{align}
\centering
	J &= \frac{1}{2}E({\|\overrightarrow{x}-W_1^T\overrightarrow{z_1}\|}^2)+L(f(x),y)
	\label{eq:sup-Lmser-objective}
\end{align}
where the additional term $L(f(x),y)$ measures the distance between the predict of Lmser net $f(x)$ with given label $y$.

\section*{Experiments}

In this section, we demonstrate the effectiveness and strengths of the deep Lmser learning by some promising results on image recognition, reconstruction, generation, and associative recall. We use Lmser(un) to denote the Lmser network trained only in the unsupervised manner, use Lmser(un-n) to denote the Lmser(un) without symmetrically sharing neurons, use Lmser(sup) to denote the Lmser network trained jointly in the supervised and unsupervised manner.

\subsection*{Datasets and experimental settings}

We evaluate Lmser on three benchmark datasests: MNIST\cite{lecun2010mnist}, Fashion-MNIST \cite{Fashion-MNIST}, and CelebA\cite{yang2015facial}.
\begin{itemize}
    \item The MNIST constains $60,000$ training images and $10,000$ test images. The images are handwritten digits from $250$ people. Each picture in the data set consists of $28 \times 28$ pixels, each of which is represented by a gray value.
    \item Fashion-MNIST is a MNIST-like dataset which shares the same image size and structure of training and testing splits. Fashion-MNIST is served as a replacement for the original MNIST dataset for benchmarking machine learning algorithms.
    \item The CelebA contains $202,599$ face images with coarse alignment \cite{yang2015facial}. The images are cropped at the center to $128\times 128$, which contain faces from various view-points and with different expressions.
\end{itemize}

For MNIST and Fashion-MNIST, we use $4$ layers with the numbers of neurons $[784,300,100,10]$ for Lmser based on fully-connected layers, and use $4$ layers with the channel sizes $[16,32,64,64]$ for the CNN based Lmser. For CelebA, we use $7$ layers with channel sizes $[16,32,64,128,256,256,256]$ for the CNN based Lmser. We train every network model with Adam (Adaptive moment estimation) optimiser method \cite{Adam2014}. In the training process, the mini-batch is set to be $50$ and the adaptive learning is set to be $0.01$ with decay rate $0.9999$.

\subsection*{Reconstruction}

\begin{figure}
	\centering
	\includegraphics[width=8cm]{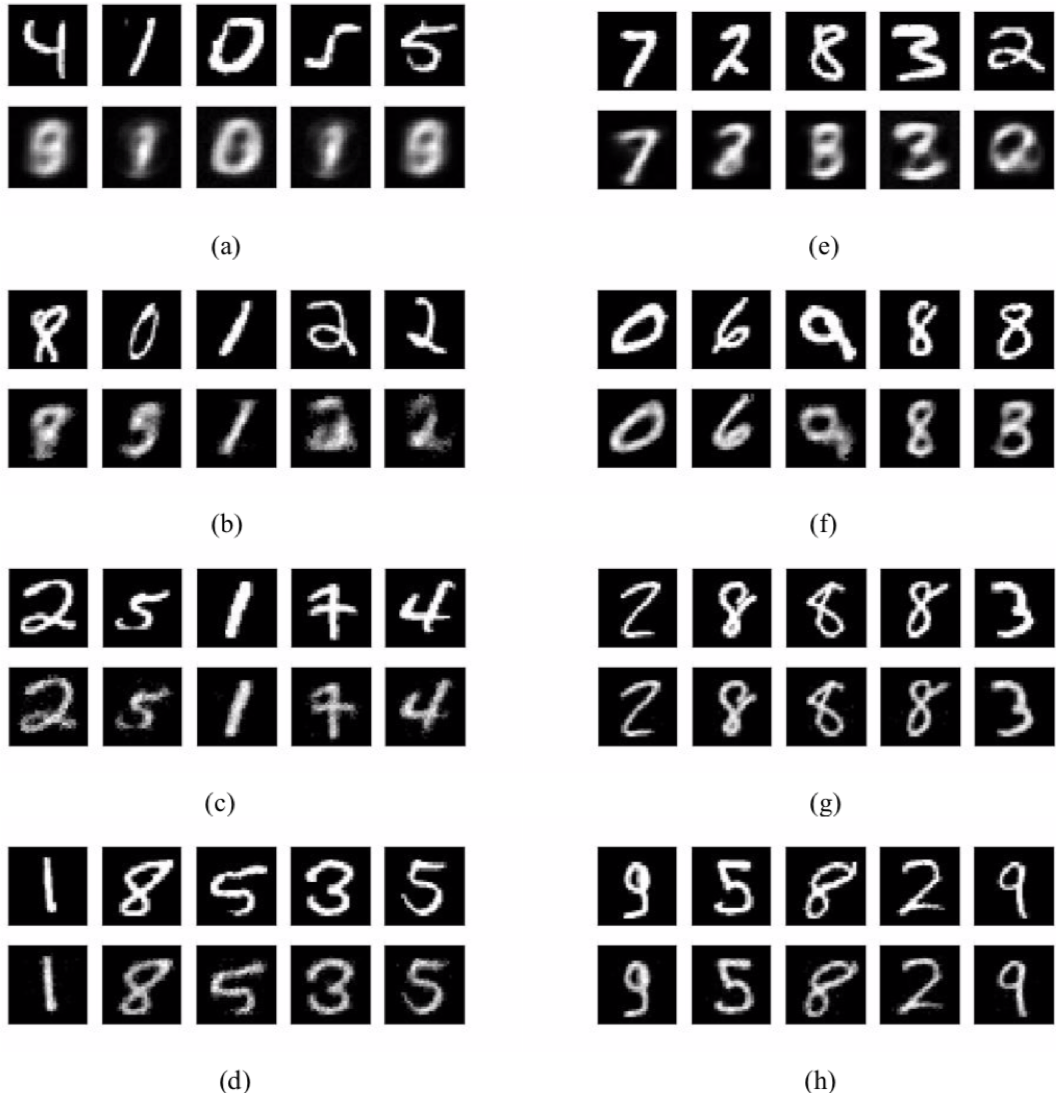}
	\caption{Examples of reconstructed images on MNIST. (a) AE, $\tau=500$; (b) Lmser(un-n), $\tau=500$; (c) Lmser(un), $\tau=500$; (d) Lmser(sup), $\tau=500$; (e) AE, $\tau=5000$; (f) Lmser(un-n), $\tau=5000$; (g) Lmser(un), $\tau=5000$; (h) Lmser(sup), $\tau=5000$.
	Notations $\tau$: the number of training iterations;	AE: autoencoder; Lmser(un): Lmser in unsupervised manner; Lmser(un-n): Lmser(un) without symmetrically sharing neurons; Lmser(sup): Lmser jointly in supervised and unsupervised manner.}
	\label{fig:recon-mnist}
\end{figure}

\begin{figure}
	\centering
	\includegraphics[width=8cm]{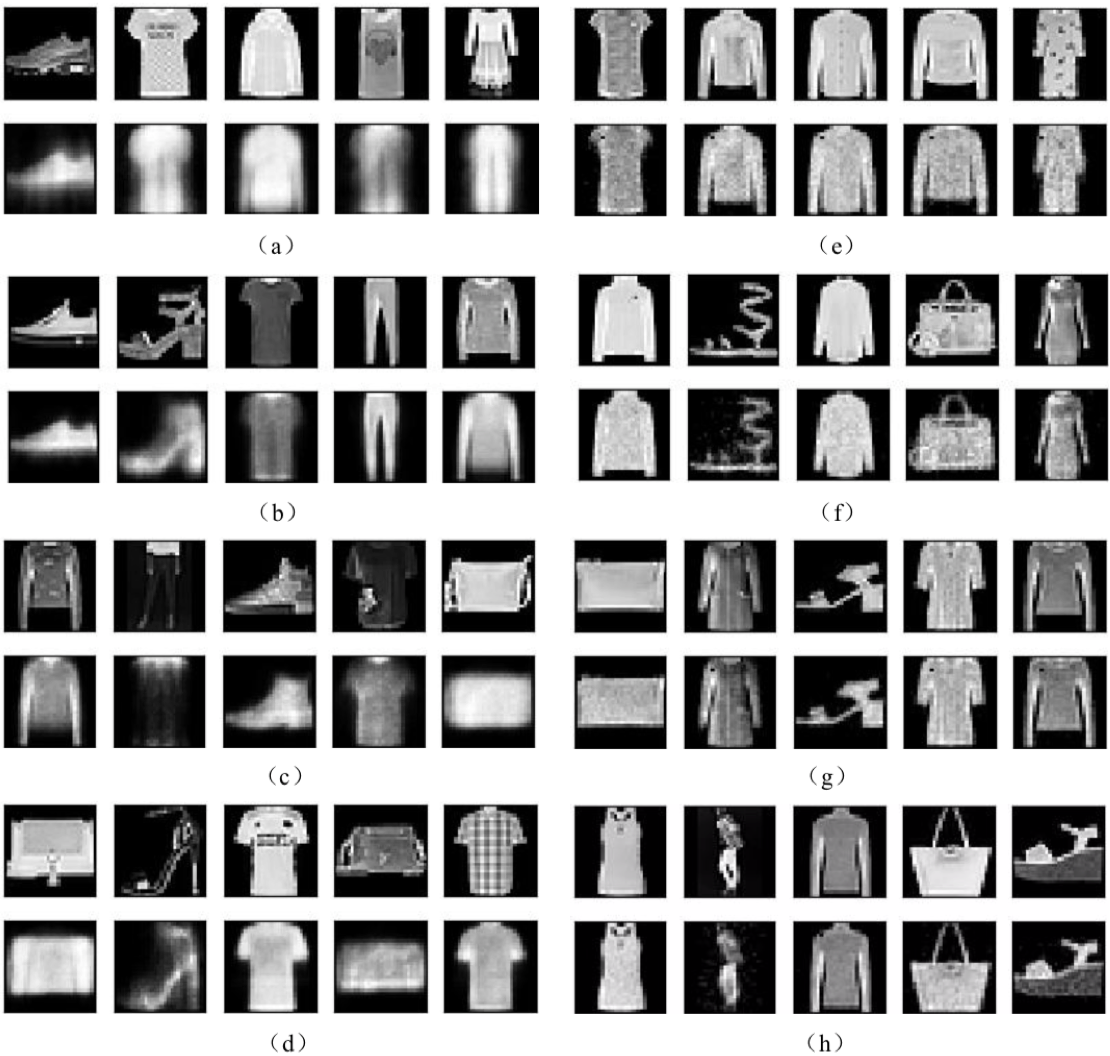}
	\caption{Examples of reconstructed images on Fashion-MNIST. (a-d) AE, $\tau=1000,5000,10000,20000$; (e-h) Lmser(un), $\tau=1000,5000,10000,20000$;
	Notations $\tau$: the number of training iterations; AE: autoencoder; Lmser(un): Lmser in unsupervised manner.}
	\label{fig:recon-fashioin-mnist}
\end{figure}

We evaluate the reconstruction performance on MNIST dataset. For comparisons, AE is used as a baseline. 
Examples of the reconstructed results are given in Figure \ref{fig:recon-mnist} for different number of training iterations. Detailed reconstruction errors are summarized in Table \ref{tab:mse-mnist}. It can be observed that Lmser converges faster with smaller reconstruction errors than AE, and Lmser(un) is better than Lmser(un-n), which suggests that shortcuts between paired neurons play a significant role in reconstruction performance. Lmser(sup) is slightly worse than Lmser(un), but still much better Lmser(un-n) and AE, indicating that there is a trade-off between the classification and reconstruction. Moreover, we also test the reconstruction performance on Fashion-MNIST. Examples are shown in Figure \ref{fig:recon-fashioin-mnist}. Obviously, Lmser converges faster and is better than AE, which is consistent to the observations from Figure \ref{fig:recon-mnist}.

\begin{table}[ht]
	\centering
	\begin{tabular}{cccc}
		\hline
		Model & $\tau=500$ & $\tau=5000$ & $\tau=20000$ \\ \hline
		AE    & 0.071      & 0.036  & 0.020  \\
		Lmser(un-n) & 0.030   & 0.021  & 0.016  \\
		Lmser(un)   & 0.0067  & 0.0018 & 0.0006 \\
		Lmser(sup)  & 0.0078  & 0.0025 & 0.0007 \\
		\hline
	\end{tabular}
	\caption{Reconstruction error on MNIST, where $\tau$ denotes the number of training iterations.  }
	\label{tab:mse-mnist}
\end{table}

\subsection*{Recognition}

In this section, we investigate the classification performance of Lmser(sup), which is trained jointly in supervised manner and unsupervised manner. The classification accuracies of Lmser(sup) and a fully-connected feedforward network (FCN) are comparably high, both over $98\%$. When there exist adversarial attacks at intensity level $0.3$ from the Fast Gradient Sign Method (FGSM) \cite{fgsm2015}, a well-known adversarial attack method. The classification accuracy of FCN drops down to $2.68\%$, while Lmser(sup) gets $31.16\%$. For convolutional layers based networks, CNN classifies MNIST with accuracy $99.48\%$ higher than FCN, and drops down to $5.2\%$ at the presence of FGSM attacks at intensity level $0.3$, while CNN based Lmser achieves still $47.9\%$, more robust than CNN.

\subsection*{Generation}

\begin{figure}[ht]
	\centering
	\subfigure[]{
		\includegraphics[width=0.45\textwidth]{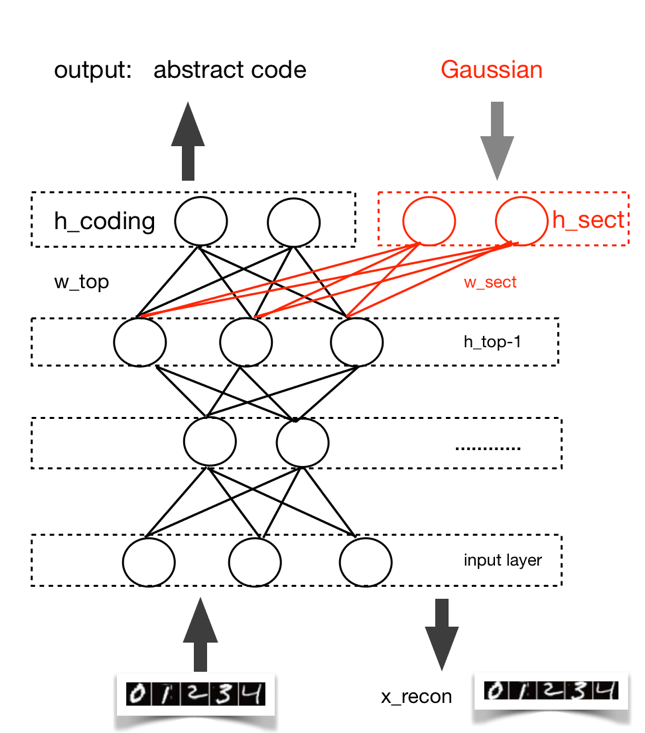}
	}	
	\subfigure[]{
		\includegraphics[width=0.45\textwidth]{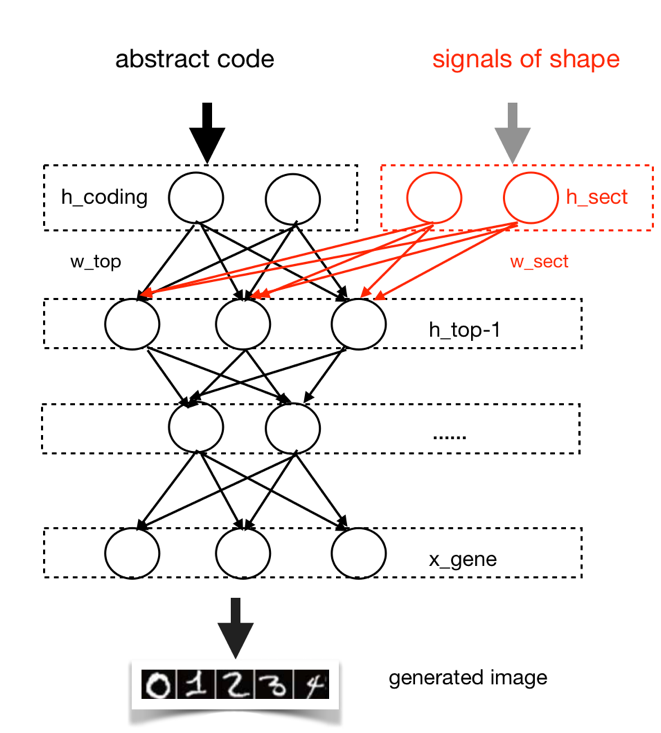}
	}
	\caption{Structure of supervised Lmser for generation: (a) training stage; (b) generation stage}
	\label{fig:structure-of-generation}
\end{figure}

\begin{figure}[ht]
	\centering
	\subfigure[]{
		\includegraphics[width=0.45\textwidth]{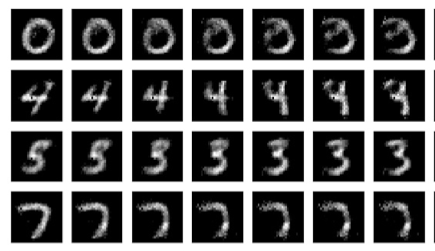}
	}	
	\subfigure[]{
		\includegraphics[width=0.45\textwidth]{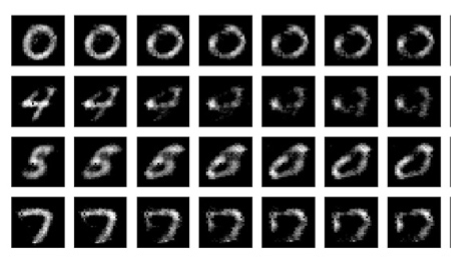}
	}
	\subfigure[]{
		\includegraphics[width=0.45\textwidth]{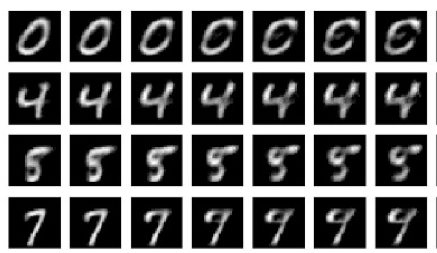}
	}
	\subfigure[]{
		\includegraphics[width=0.45\textwidth]{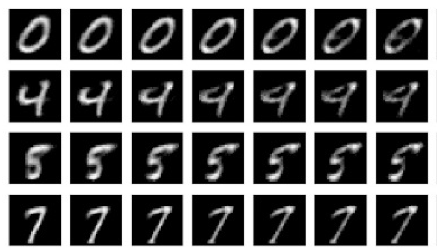}
	}
	\caption{Generated digital numbers by (a-b) Lmser(sup); (c-d) Lmser(un). There are $10$ codes in the latent space, where $Z_i$ represent the $i$-th coding unit. (a) manipulating $Z_3$; (b) manipulating $Z_0$; (c) manipulating $Z_5$; (d) manipulating $Z_9$.}
	\label{fig:fcn-generation}
\end{figure}

In this section, we investigate the performance of deep Lmser learning in image generation by manipulating the latent code to get different styles of digital numbers and clothes. As shown in Figure \ref{fig:structure-of-generation}, to get different styles of digital numbers, we add two more hidden coding units to the top layer of Lmser(sup), except for the category coding units, to capture the non-categorical information. We assume that the two additional coding units are independent and follow Gaussian distributions. In practice, when we train the model Lmser(sup), the means of the two additional coding units are computed by encoder. Before they are fed into the decoder, we add a Gaussian noise to them. When generating digital numbers, as shown in Figure \ref{fig:structure-of-generation} (b), the decoder will generate digital numbers according to the label information by the category coding units and random style signals by the two additional hidden coding units. Figure \ref{fig:fcn-generation}(a-b) show that when assigning the coding unit with different values, the digital number is gradually changing.

For the input patterns without labels, no labels can be used to guide the separation of categorical information and non-categorical styles. Lmser is still able to self-organize the input patterns $X$ in the top coding domain $Y$, according to the similarities in semantics, topology and styles. By manipulating one coding unit in the top layer with others fixed, we can observe pattern changing smoothly as the coding value varies, which indicates that the coding regions are well self-organized and clustered. With such property, we are able to synthesize images in a controllable way or in a creative way for novel synthesis and reasoning. Figure \ref{fig:fcn-generation}(c) shows that when specifying the coding unit with values from $0$ to $10$, the digital number is gradually changed to have a circle created over the head, while in Figure \ref{fig:fcn-generation}(d) another coding unit seems to control the tilt degree of generated images.

\begin{figure}[ht]
	\centering
	\subfigure[]{
		\includegraphics[width=0.45\textwidth]{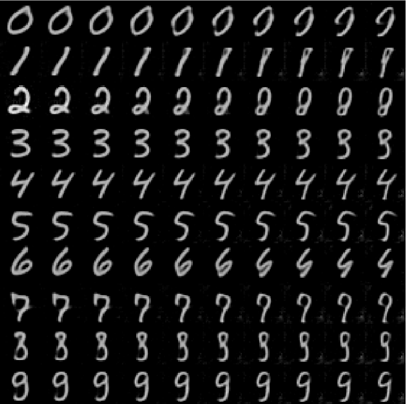}
	}	
	\subfigure[]{
		\includegraphics[width=0.45\textwidth]{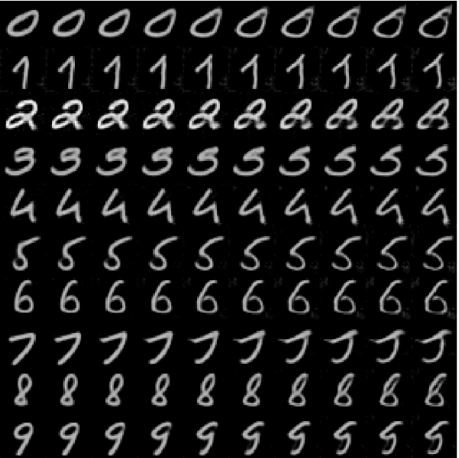}
	}
	\subfigure[]{
		\includegraphics[width=0.6\textwidth]{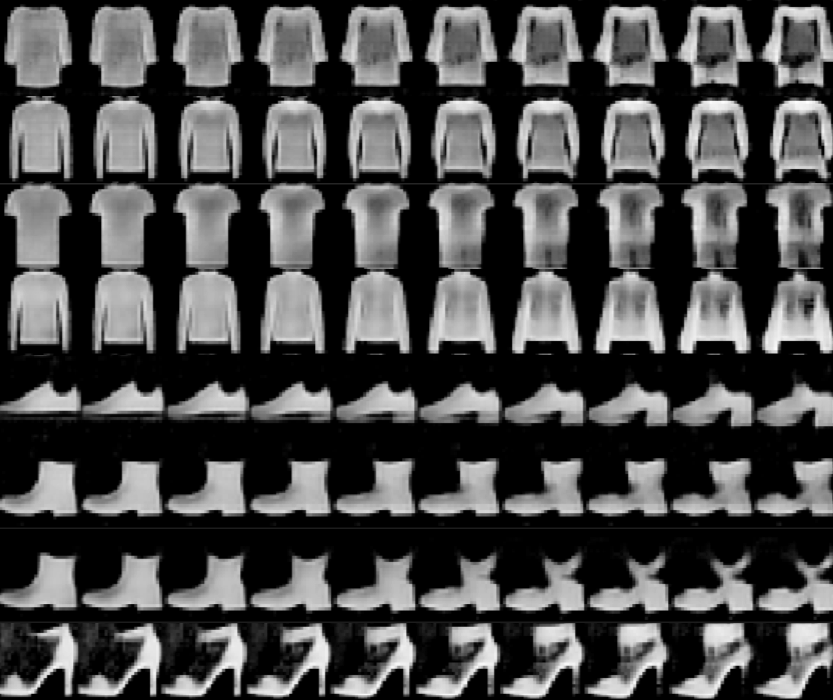}
	}
	\caption{Image generated by cnn based Lmser in unsupervised learning manner. (a-b) on MNIST; (c) on Fashion-MNIST.}
	\label{fig:cnn-generation}
\end{figure}

We also test the generation performance of CNN based Lmser network, and examples are shown in Figure \ref{fig:cnn-generation}. The pattern changing is similar to that in Figure \ref{fig:fcn-generation} by the Lmser based the fully-connected layers, but the quality of the synthesized images is much higher, possibly due to the advantages of the convolutional layers in extracting and generating image neighborhood patterns. In Figure \ref{fig:fcn-generation}(a-b), we can see clearly that some novel digits are generated, and they look plausible because their structures resemble the original digits. In Figure \ref{fig:fcn-generation}(c), new styles of clothes or shoes are generated, which shows the potential application of Lmser on creative design.

\subsection*{Association}

For the incomplete input, Lmser is able to recover the missing part by associative memory from the observed part. This function is related to tasks such as associative recall of someone’s whole face from a face image with key parts blocked. Specifically, the observed part of the image is fed into the Lmser network, and triggers the bottom-up signals passing to the top layers. Then, the activated neurons passing the top-down signals back to the bottom layer to give a complete image, which actually recovers unobserved part based on what has been learned by the network layers. The sharing connection weights by both directions of the links between the consecutive layers enables Lmser to catch invertible structures under input patterns for restoring these structures under partial input.

\begin{figure}[ht]
	\centering
	\subfigure[masked on eyes]{
		\includegraphics[width=0.8\textwidth]{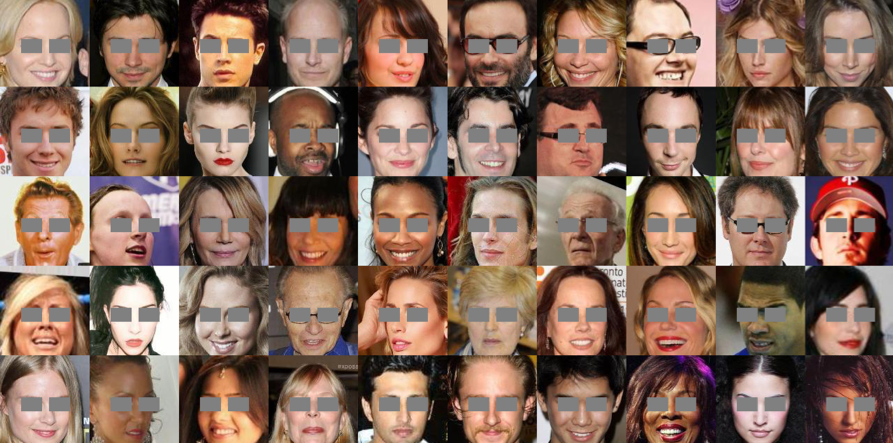}
	}
	\subfigure[association]{
		\includegraphics[width=0.8\textwidth]{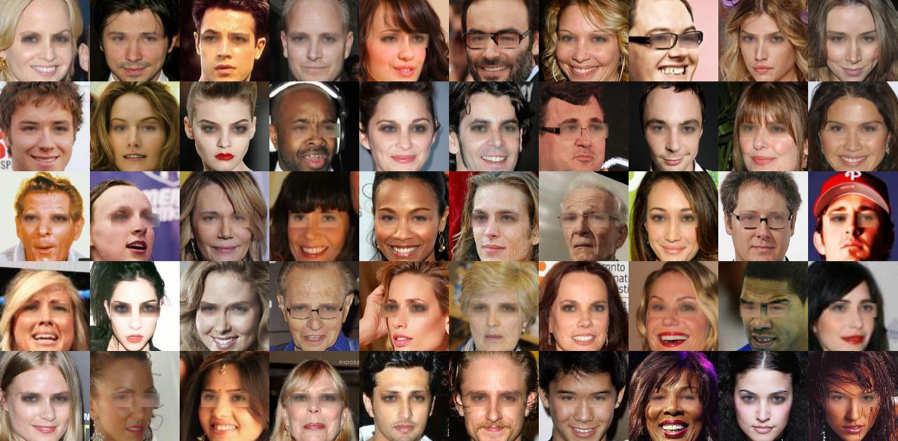}
	}
	\subfigure[ground truth]{
		\includegraphics[width=0.8\textwidth]{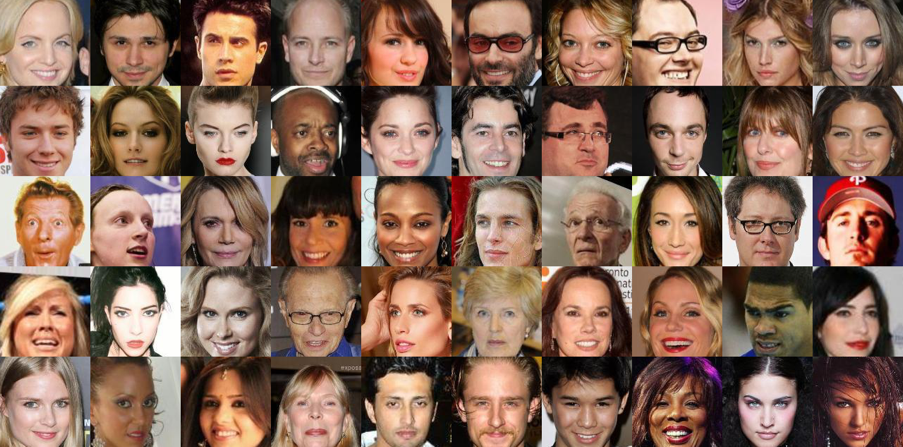}
	}	

	\caption{Examples of associative recall from partial face images. (a) partial face images with eyes blocked; (b) associative recall by CNN based Lmser(un-n); (c) the ground-truth images. }
	\label{fig:asso}
\end{figure}

We train the CNN based Lmser net on the CelebA dataset, and then feed the masked images into the model for the output of reconstructed complete images. In practice, we have found that shortcuts between paired neurons is helpful in reconstruction, but not so beneficial in associative memory. Therefore, we only present the results by the CNN based Lmser net without using the paired neurons for both encoder and decoder in Figure \ref{fig:asso}. Examples of the face images with eyes blocked are given in Figure \ref{fig:asso}(a), and the corresponding reconstructed complete face from the associative memory by Lmser are shown in Figure \ref{fig:asso}(b), which are comparable to the ground-truth images in Figure \ref{fig:asso}. The results demonstrate that Lmser is promising for this task of associative recall from partial input.

\section*{Conclusion}

In this paper, we have revisited the Lmser network for the practical implementation of multiple layers, and confirmed that several of its potential functions indeed work effectively via experiments on image recognition, reconstruction and associative memory, and generation. Furthermore, we extend to build the Lmser network based on convolutional layers. Experiments demonstrate that Lmser not only works as preliminarily discussed in the original paper, but also is promising in various applications.

\begin{backmatter}

\section*{Competing interests}
  The authors declare that they have no competing interests.

\section*{Ethics approval and consent to participate}
Not applicable.

\section*{Consent for publication}
All authors agree to publish this paper.

\section*{Availability of data and material}
The data involved in experiments are all public data from public web sites.

\section*{Funding}
WF220103010 from Shanghai Jiao Tong University;

\section*{Authors' contributions}
Lei Xu conceived the whole project. Wenjing Huang and Shikui Tu wrote the paper. Wenjing Huang perform the analysis.


\section*{Acknowledgements}
This work was supported by the Zhi-Yuan chair professorship start-up grant (WF220103010) from Shanghai Jiao Tong University. Wenjing Huang would thank Yuze Guo for the help in image reconstruction experiments.


\bibliographystyle{bmc-mathphys} 
\bibliography{ref}


\begin{thebibliography}{18}
\ifx \bisbn   \undefined \def \bisbn  #1{ISBN #1}\fi
\ifx \binits  \undefined \def \binits#1{#1}\fi
\ifx \bauthor  \undefined \def \bauthor#1{#1}\fi
\ifx \batitle  \undefined \def \batitle#1{#1}\fi
\ifx \bjtitle  \undefined \def \bjtitle#1{#1}\fi
\ifx \bvolume  \undefined \def \bvolume#1{\textbf{#1}}\fi
\ifx \byear  \undefined \def \byear#1{#1}\fi
\ifx \bissue  \undefined \def \bissue#1{#1}\fi
\ifx \bfpage  \undefined \def \bfpage#1{#1}\fi
\ifx \blpage  \undefined \def \blpage #1{#1}\fi
\ifx \burl  \undefined \def \burl#1{\textsf{#1}}\fi
\ifx \doiurl  \undefined \def \doiurl#1{\textsf{#1}}\fi
\ifx \betal  \undefined \def \betal{\textit{et al.}}\fi
\ifx \binstitute  \undefined \def \binstitute#1{#1}\fi
\ifx \binstitutionaled  \undefined \def \binstitutionaled#1{#1}\fi
\ifx \bctitle  \undefined \def \bctitle#1{#1}\fi
\ifx \beditor  \undefined \def \beditor#1{#1}\fi
\ifx \bpublisher  \undefined \def \bpublisher#1{#1}\fi
\ifx \bbtitle  \undefined \def \bbtitle#1{#1}\fi
\ifx \bedition  \undefined \def \bedition#1{#1}\fi
\ifx \bseriesno  \undefined \def \bseriesno#1{#1}\fi
\ifx \blocation  \undefined \def \blocation#1{#1}\fi
\ifx \bsertitle  \undefined \def \bsertitle#1{#1}\fi
\ifx \bsnm \undefined \def \bsnm#1{#1}\fi
\ifx \bsuffix \undefined \def \bsuffix#1{#1}\fi
\ifx \bparticle \undefined \def \bparticle#1{#1}\fi
\ifx \barticle \undefined \def \barticle#1{#1}\fi
\ifx \bconfdate \undefined \def \bconfdate #1{#1}\fi
\ifx \botherref \undefined \def \botherref #1{#1}\fi
\ifx \url \undefined \def \url#1{\textsf{#1}}\fi
\ifx \bchapter \undefined \def \bchapter#1{#1}\fi
\ifx \bbook \undefined \def \bbook#1{#1}\fi
\ifx \bcomment \undefined \def \bcomment#1{#1}\fi
\ifx \oauthor \undefined \def \oauthor#1{#1}\fi
\ifx \citeauthoryear \undefined \def \citeauthoryear#1{#1}\fi
\ifx \endbibitem  \undefined \def \endbibitem {}\fi
\ifx \bconflocation  \undefined \def \bconflocation#1{#1}\fi
\ifx \arxivurl  \undefined \def \arxivurl#1{\textsf{#1}}\fi
\csname PreBibitemsHook\endcsname

\bibitem{Ballard1987}
\begin{bchapter}
\bauthor{\bsnm{Ballard}, \binits{D.H.}}:
\bctitle{Modular learning in neural networks}.
In: \bbtitle{Proceedings of the Sixth National Conference on Artificial
  Intelligence - Volume 1}.
\bsertitle{AAAI'87},
pp. \bfpage{279}--\blpage{284}
(\byear{1987})
\end{bchapter}
\endbibitem

\bibitem{Bourlard1988AE}
\begin{barticle}
\bauthor{\bsnm{Bourlard}, \binits{H.}},
\bauthor{\bsnm{Kamp}, \binits{Y.}}:
\batitle{Auto-association by multilayer perceptrons and singular value
  decomposition}.
\bjtitle{Biol. Cybern.}
\bvolume{59}(\bissue{4-5}),
\bfpage{291}--\blpage{294}
(\byear{1988})
\end{barticle}
\endbibitem

\bibitem{xu1991LMSER}
\begin{bchapter}
\bauthor{\bsnm{Xu}, \binits{L.}}:
\bctitle{Least mse reconstruction for self-organization: (i)\&(ii)}.
In: \bbtitle{Proc. of 1991 International Joint Conference on Neural Networks},
pp. \bfpage{2363}--\blpage{2373}
(\byear{1991})
\end{bchapter}
\endbibitem

\bibitem{xu1993least}
\begin{barticle}
\bauthor{\bsnm{Xu}, \binits{L.}}:
\batitle{Least mean square error reconstruction principle for self-organizing
  neural-nets}.
\bjtitle{Neural networks}
\bvolume{6}(\bissue{5}),
\bfpage{627}--\blpage{648}
(\byear{1993})
\end{barticle}
\endbibitem

\bibitem{hinton2006fast}
\begin{barticle}
\bauthor{\bsnm{Hinton}, \binits{G.E.}},
\bauthor{\bsnm{Osindero}, \binits{S.}},
\bauthor{\bsnm{Teh}, \binits{Y.-W.}}:
\batitle{A fast learning algorithm for deep belief nets}.
\bjtitle{Neural computation}
\bvolume{18}(\bissue{7}),
\bfpage{1527}--\blpage{1554}
(\byear{2006})
\end{barticle}
\endbibitem

\bibitem{ronneberger2015u}
\begin{bchapter}
\bauthor{\bsnm{Ronneberger}, \binits{O.}},
\bauthor{\bsnm{Fischer}, \binits{P.}},
\bauthor{\bsnm{Brox}, \binits{T.}}:
\bctitle{U-net: Convolutional networks for biomedical image segmentation}.
In: \bbtitle{International Conference on Medical Image Computing and
  Computer-assisted Intervention},
pp. \bfpage{234}--\blpage{241}
(\byear{2015}).
\bcomment{Springer}
\end{bchapter}
\endbibitem

\bibitem{mao2016image}
\begin{bchapter}
\bauthor{\bsnm{Mao}, \binits{X.}},
\bauthor{\bsnm{Shen}, \binits{C.}},
\bauthor{\bsnm{Yang}, \binits{Y.-B.}}:
\bctitle{Image restoration using very deep convolutional encoder-decoder
  networks with symmetric skip connections}.
In: \bbtitle{Advances in Neural Information Processing Systems},
pp. \bfpage{2802}--\blpage{2810}
(\byear{2016})
\end{bchapter}
\endbibitem

\bibitem{kingma2013auto}
\begin{botherref}
\oauthor{\bsnm{Kingma}, \binits{D.P.}},
\oauthor{\bsnm{Welling}, \binits{M.}}:
Auto-encoding variational bayes.
arXiv preprint arXiv:1312.6114
(2013)
\end{botherref}
\endbibitem

\bibitem{goodfellow2014generative}
\begin{bchapter}
\bauthor{\bsnm{Goodfellow}, \binits{I.}},
\bauthor{\bsnm{Pouget-Abadie}, \binits{J.}},
\bauthor{\bsnm{Mirza}, \binits{M.}},
\bauthor{\bsnm{Xu}, \binits{B.}},
\bauthor{\bsnm{Warde-Farley}, \binits{D.}},
\bauthor{\bsnm{Ozair}, \binits{S.}},
\bauthor{\bsnm{Courville}, \binits{A.}},
\bauthor{\bsnm{Bengio}, \binits{Y.}}:
\bctitle{Generative adversarial nets}.
In: \bbtitle{Advances in Neural Information Processing Systems},
pp. \bfpage{2672}--\blpage{2680}
(\byear{2014})
\end{bchapter}
\endbibitem

\bibitem{lecun1989backpropagation}
\begin{barticle}
\bauthor{\bsnm{LeCun}, \binits{Y.}},
\bauthor{\bsnm{Boser}, \binits{B.}},
\bauthor{\bsnm{Denker}, \binits{J.S.}},
\bauthor{\bsnm{Henderson}, \binits{D.}},
\bauthor{\bsnm{Howard}, \binits{R.E.}},
\bauthor{\bsnm{Hubbard}, \binits{W.}},
\bauthor{\bsnm{Jackel}, \binits{L.D.}}:
\batitle{Backpropagation applied to handwritten zip code recognition}.
\bjtitle{Neural computation}
\bvolume{1}(\bissue{4}),
\bfpage{541}--\blpage{551}
(\byear{1989})
\end{barticle}
\endbibitem

\bibitem{hinton2006reducing}
\begin{barticle}
\bauthor{\bsnm{Hinton}, \binits{G.E.}},
\bauthor{\bsnm{Salakhutdinov}, \binits{R.R.}}:
\batitle{Reducing the dimensionality of data with neural networks}.
\bjtitle{science}
\bvolume{313}(\bissue{5786}),
\bfpage{504}--\blpage{507}
(\byear{2006})
\end{barticle}
\endbibitem

\bibitem{he2016deep}
\begin{bchapter}
\bauthor{\bsnm{He}, \binits{K.}},
\bauthor{\bsnm{Zhang}, \binits{X.}},
\bauthor{\bsnm{Ren}, \binits{S.}},
\bauthor{\bsnm{Sun}, \binits{J.}}:
\bctitle{Deep residual learning for image recognition}.
In: \bbtitle{Proceedings of the IEEE Conference on Computer Vision and Pattern
  Recognition},
pp. \bfpage{770}--\blpage{778}
(\byear{2016})
\end{bchapter}
\endbibitem

\bibitem{DenseNet2016}
\begin{botherref}
\oauthor{\bsnm{Huang}, \binits{G.}},
\oauthor{\bsnm{Liu}, \binits{Z.}},
\oauthor{\bsnm{Weinberger}, \binits{K.Q.}}:
Densely connected convolutional networks.
CoRR
\textbf{abs/1608.06993}
(2016).
\arxivurl{1608.06993}
\end{botherref}
\endbibitem

\bibitem{lecun2010mnist}
\begin{botherref}
\oauthor{\bsnm{LeCun}, \binits{Y.}},
\oauthor{\bsnm{Cortes}, \binits{C.}},
\oauthor{\bsnm{Burges}, \binits{C.}}:
Mnist handwritten digit database.
AT\&T Labs [Online]. Available: http://yann. lecun. com/exdb/mnist
\textbf{2}
(2010)
\end{botherref}
\endbibitem

\bibitem{Fashion-MNIST}
\begin{botherref}
\oauthor{\bsnm{Xiao}, \binits{H.}},
\oauthor{\bsnm{Rasul}, \binits{K.}},
\oauthor{\bsnm{Vollgraf}, \binits{R.}}:
Fashion-MNIST: a Novel Image Dataset for Benchmarking Machine Learning
  Algorithms.
\arxivurl{cs.LG/1708.07747}
\end{botherref}
\endbibitem

\bibitem{yang2015facial}
\begin{bchapter}
\bauthor{\bsnm{Yang}, \binits{S.}},
\bauthor{\bsnm{Luo}, \binits{P.}},
\bauthor{\bsnm{Loy}, \binits{C.-C.}},
\bauthor{\bsnm{Tang}, \binits{X.}}:
\bctitle{From facial parts responses to face detection: A deep learning
  approach}.
In: \bbtitle{Proceedings of the IEEE International Conference on Computer
  Vision},
pp. \bfpage{3676}--\blpage{3684}
(\byear{2015})
\end{bchapter}
\endbibitem

\bibitem{Adam2014}
\begin{botherref}
\oauthor{\bsnm{Kingma}, \binits{D.P.}},
\oauthor{\bsnm{Ba}, \binits{J.}}:
Adam: {A} method for stochastic optimization.
CoRR
\textbf{abs/1412.6980}
(2014).
\arxivurl{1412.6980}
\end{botherref}
\endbibitem

\bibitem{fgsm2015}
\begin{bchapter}
\bauthor{\bsnm{Goodfellow}, \binits{I.}},
\bauthor{\bsnm{Shlens}, \binits{J.}},
\bauthor{\bsnm{Szegedy}, \binits{C.}}:
\bctitle{Explaining and harnessing adversarial examples}.
In: \bbtitle{International Conference on Learning Representations}
(\byear{2015}).
\burl{http://arxiv.org/abs/1412.6572}
\end{bchapter}
\endbibitem

\end{thebibliography}

\newcommand{\BMCxmlcomment}[1]{}

\BMCxmlcomment{

<refgrp>

<bibl id="B1">
  <title><p>Modular Learning in Neural Networks</p></title>
  <aug>
    <au><snm>Ballard</snm><fnm>DH</fnm></au>
  </aug>
  <source>Proceedings of the Sixth National Conference on Artificial
  Intelligence - Volume 1</source>
  <series><title><p>AAAI'87</p></title></series>
  <pubdate>1987</pubdate>
  <fpage>279</fpage>
  <lpage>-284</lpage>
</bibl>

<bibl id="B2">
  <title><p>Auto-association by Multilayer Perceptrons and Singular Value
  Decomposition</p></title>
  <aug>
    <au><snm>Bourlard</snm><fnm>H.</fnm></au>
    <au><snm>Kamp</snm><fnm>Y.</fnm></au>
  </aug>
  <source>Biol. Cybern.</source>
  <publisher>Secaucus, NJ, USA: Springer-Verlag New York, Inc.</publisher>
  <pubdate>1988</pubdate>
  <volume>59</volume>
  <issue>4-5</issue>
  <fpage>291</fpage>
  <lpage>-294</lpage>
</bibl>

<bibl id="B3">
  <title><p>Least MSE Reconstruction for Self-Organization:
  (I)&amp(II)</p></title>
  <aug>
    <au><snm>Xu</snm><fnm>L</fnm></au>
  </aug>
  <source>Proc. of 1991 International Joint Conference on Neural
  Networks</source>
  <pubdate>1991</pubdate>
  <fpage>2363</fpage>
  <lpage>-2373</lpage>
</bibl>

<bibl id="B4">
  <title><p>Least mean square error reconstruction principle for
  self-organizing neural-nets</p></title>
  <aug>
    <au><snm>Xu</snm><fnm>L</fnm></au>
  </aug>
  <source>Neural networks</source>
  <pubdate>1993</pubdate>
  <volume>6</volume>
  <issue>5</issue>
  <fpage>627</fpage>
  <lpage>-648</lpage>
</bibl>

<bibl id="B5">
  <title><p>A fast learning algorithm for deep belief nets</p></title>
  <aug>
    <au><snm>Hinton</snm><fnm>GE</fnm></au>
    <au><snm>Osindero</snm><fnm>S</fnm></au>
    <au><snm>Teh</snm><fnm>YW</fnm></au>
  </aug>
  <source>Neural computation</source>
  <publisher>MIT Press</publisher>
  <pubdate>2006</pubdate>
  <volume>18</volume>
  <issue>7</issue>
  <fpage>1527</fpage>
  <lpage>-1554</lpage>
</bibl>

<bibl id="B6">
  <title><p>U-net: Convolutional networks for biomedical image
  segmentation</p></title>
  <aug>
    <au><snm>Ronneberger</snm><fnm>O</fnm></au>
    <au><snm>Fischer</snm><fnm>P</fnm></au>
    <au><snm>Brox</snm><fnm>T</fnm></au>
  </aug>
  <source>International Conference on Medical image computing and
  computer-assisted intervention</source>
  <pubdate>2015</pubdate>
  <fpage>234</fpage>
  <lpage>-241</lpage>
</bibl>

<bibl id="B7">
  <title><p>Image restoration using very deep convolutional encoder-decoder
  networks with symmetric skip connections</p></title>
  <aug>
    <au><snm>Mao</snm><fnm>X</fnm></au>
    <au><snm>Shen</snm><fnm>C</fnm></au>
    <au><snm>Yang</snm><fnm>YB</fnm></au>
  </aug>
  <source>Advances in neural information processing systems</source>
  <pubdate>2016</pubdate>
  <fpage>2802</fpage>
  <lpage>-2810</lpage>
</bibl>

<bibl id="B8">
  <title><p>Auto-encoding variational bayes</p></title>
  <aug>
    <au><snm>Kingma</snm><fnm>DP</fnm></au>
    <au><snm>Welling</snm><fnm>M</fnm></au>
  </aug>
  <source>arXiv preprint arXiv:1312.6114</source>
  <pubdate>2013</pubdate>
</bibl>

<bibl id="B9">
  <title><p>Generative adversarial nets</p></title>
  <aug>
    <au><snm>Goodfellow</snm><fnm>I</fnm></au>
    <au><snm>Pouget Abadie</snm><fnm>J</fnm></au>
    <au><snm>Mirza</snm><fnm>M</fnm></au>
    <au><snm>Xu</snm><fnm>B</fnm></au>
    <au><snm>Warde Farley</snm><fnm>D</fnm></au>
    <au><snm>Ozair</snm><fnm>S</fnm></au>
    <au><snm>Courville</snm><fnm>A</fnm></au>
    <au><snm>Bengio</snm><fnm>Y</fnm></au>
  </aug>
  <source>Advances in neural information processing systems</source>
  <pubdate>2014</pubdate>
  <fpage>2672</fpage>
  <lpage>-2680</lpage>
</bibl>

<bibl id="B10">
  <title><p>Backpropagation applied to handwritten zip code
  recognition</p></title>
  <aug>
    <au><snm>LeCun</snm><fnm>Y</fnm></au>
    <au><snm>Boser</snm><fnm>B</fnm></au>
    <au><snm>Denker</snm><fnm>JS</fnm></au>
    <au><snm>Henderson</snm><fnm>D</fnm></au>
    <au><snm>Howard</snm><fnm>RE</fnm></au>
    <au><snm>Hubbard</snm><fnm>W</fnm></au>
    <au><snm>Jackel</snm><fnm>LD</fnm></au>
  </aug>
  <source>Neural computation</source>
  <publisher>MIT Press</publisher>
  <pubdate>1989</pubdate>
  <volume>1</volume>
  <issue>4</issue>
  <fpage>541</fpage>
  <lpage>-551</lpage>
</bibl>

<bibl id="B11">
  <title><p>Reducing the dimensionality of data with neural
  networks</p></title>
  <aug>
    <au><snm>Hinton</snm><fnm>GE</fnm></au>
    <au><snm>Salakhutdinov</snm><fnm>RR</fnm></au>
  </aug>
  <source>science</source>
  <publisher>American Association for the Advancement of Science</publisher>
  <pubdate>2006</pubdate>
  <volume>313</volume>
  <issue>5786</issue>
  <fpage>504</fpage>
  <lpage>-507</lpage>
</bibl>

<bibl id="B12">
  <title><p>Deep residual learning for image recognition</p></title>
  <aug>
    <au><snm>He</snm><fnm>K</fnm></au>
    <au><snm>Zhang</snm><fnm>X</fnm></au>
    <au><snm>Ren</snm><fnm>S</fnm></au>
    <au><snm>Sun</snm><fnm>J</fnm></au>
  </aug>
  <source>Proceedings of the IEEE conference on computer vision and pattern
  recognition</source>
  <pubdate>2016</pubdate>
  <fpage>770</fpage>
  <lpage>-778</lpage>
</bibl>

<bibl id="B13">
  <title><p>Densely Connected Convolutional Networks</p></title>
  <aug>
    <au><snm>Huang</snm><fnm>G</fnm></au>
    <au><snm>Liu</snm><fnm>Z</fnm></au>
    <au><snm>Weinberger</snm><fnm>KQ</fnm></au>
  </aug>
  <source>CoRR</source>
  <pubdate>2016</pubdate>
  <volume>abs/1608.06993</volume>
  <url>http://arxiv.org/abs/1608.06993</url>
</bibl>

<bibl id="B14">
  <title><p>MNIST handwritten digit database</p></title>
  <aug>
    <au><snm>LeCun</snm><fnm>Y</fnm></au>
    <au><snm>Cortes</snm><fnm>C</fnm></au>
    <au><snm>Burges</snm><fnm>CJ</fnm></au>
  </aug>
  <source>AT\&T Labs [Online]. Available: http://yann. lecun.
  com/exdb/mnist</source>
  <pubdate>2010</pubdate>
  <volume>2</volume>
</bibl>

<bibl id="B15">
  <title><p>Fashion-MNIST: a Novel Image Dataset for Benchmarking Machine
  Learning Algorithms</p></title>
  <aug>
    <au><snm>Xiao</snm><fnm>H</fnm></au>
    <au><snm>Rasul</snm><fnm>K</fnm></au>
    <au><snm>Vollgraf</snm><fnm>R</fnm></au>
  </aug>
  <pubdate>2017</pubdate>
</bibl>

<bibl id="B16">
  <title><p>From facial parts responses to face detection: A deep learning
  approach</p></title>
  <aug>
    <au><snm>Yang</snm><fnm>S</fnm></au>
    <au><snm>Luo</snm><fnm>P</fnm></au>
    <au><snm>Loy</snm><fnm>CC</fnm></au>
    <au><snm>Tang</snm><fnm>X</fnm></au>
  </aug>
  <source>Proceedings of the IEEE International Conference on Computer
  Vision</source>
  <pubdate>2015</pubdate>
  <fpage>3676</fpage>
  <lpage>-3684</lpage>
</bibl>

<bibl id="B17">
  <title><p>Adam: {A} Method for Stochastic Optimization</p></title>
  <aug>
    <au><snm>Kingma</snm><fnm>DP</fnm></au>
    <au><snm>Ba</snm><fnm>J</fnm></au>
  </aug>
  <source>CoRR</source>
  <pubdate>2014</pubdate>
  <volume>abs/1412.6980</volume>
  <url>http://arxiv.org/abs/1412.6980</url>
</bibl>

<bibl id="B18">
  <title><p>Explaining and Harnessing Adversarial Examples</p></title>
  <aug>
    <au><snm>Goodfellow</snm><fnm>I</fnm></au>
    <au><snm>Shlens</snm><fnm>J</fnm></au>
    <au><snm>Szegedy</snm><fnm>C</fnm></au>
  </aug>
  <source>International Conference on Learning Representations</source>
  <pubdate>2015</pubdate>
  <url>http://arxiv.org/abs/1412.6572</url>
</bibl>

</refgrp>
} 

\section*{Additional Files}
  \subsection*{Additional file 1 --- Sample additional file title}
    Additional file descriptions text (including details of how to
    view the file, if it is in a non-standard format or the file extension).  This might
    refer to a multi-page table or a figure.

  \subsection*{Additional file 2 --- Sample additional file title}
    Additional file descriptions text.

\end{backmatter}
\end{document}